\documentclass{article}
\usepackage{spconf}
\usepackage{graphicx}
\usepackage{amsmath}
\usepackage{amssymb}
\usepackage[colorlinks]{hyperref}
\usepackage{verbatim}
\usepackage{color}

\def\0{{\bf 0}}
\def\1{{\bf 1}}

\def\ie{{\em i.e.}}

\graphicspath{{figs/}}

\usepackage{tabularx} % added
\usepackage{booktabs}   %three lines for table, added
\usepackage{mathrsfs} %added
\usepackage{multirow} %added
\usepackage{amsfonts} %added
\usepackage{wrapfig}  %added
\usepackage[misc]{ifsym}
\usepackage{cite}
\usepackage{subcaption}
\usepackage{lipsum} % For generating sample text
\title{R2Gen-Mamba: A Selective State Space Model for Radiology Report Generation\thanks{This work has been submitted to the IEEE for possible publication. Copyright may be transferred without notice, after which this version may no longer be accessible.}}
\name{Yongheng~Sun$^{1,2}$,
Yueh Z. Lee$^{1}$, 
Genevieve A. Woodard$^{1}$, 
Hongtu Zhu$^{3}$,
Chunfeng Lian$^{2,*}$, 
Mingxia~Liu$^{1,*}$}
\address{
$^1$Department of Radiology and BRIC, UNC at Chapel Hill, Chapel Hill, NC 27599, USA\\
$^2$School of Mathematics and Statistics, Xi'an Jiaotong University, Xi'an 710049, China\\
$^3$Department of Biostatistics and BRIC, UNC at Chapel Hill, Chapel Hill, NC 27599, USA}

\begin{document}
%\ninept
%
\maketitle

\begin{abstract}
Radiology report generation is crucial in medical imaging, but the manual annotation process by physicians is time-consuming and labor-intensive, necessitating the development of automatic report generation methods.  
%While many approaches have traditionally relied on Transformer models, these can be computationally intensive, limiting their applications in real-word applications. % and less efficient.
Existing research predominantly utilizes Transformers to generate radiology reports, which can be computationally intensive, limiting their use in real applications. 
In this work, we present \emph{R2Gen-Mamba}, a novel automatic radiology report generation method that %synergistically
leverages the efficient sequence processing of the Mamba with the contextual benefits of Transformer architectures. 
Due to lower computational complexity of Mamba, R2Gen-Mamba not only enhances training and inference efficiency but also produces high-quality reports. 
Experimental results on % conducted on %the IU-X-RAY and MIMIC-CXR 
two benchmark   
datasets with more than 210,000 X-ray image-report pairs demonstrate the effectiveness of R2Gen-Mamba regarding report quality
and computational efficiency compared with several state-of-the-art methods.  
The source code can be accessed \href{https://github.com/YonghengSun1997/R2Gen-Mamba}{online}.
%and highlight its potential in advancing automated radiology report generation.
\end{abstract}

\section{INTRODUCTION}
\label{sec:intro}
%Radiology report generation is crucial in the field of medical image analysis, providing essential information for diagnosing and managing patient conditions. 
Radiology report generation is crucial in medical imaging, offering key information necessary for diagnosing and managing patient conditions. 
Traditionally, these reports are manually annotated by physicians, which is time-consuming and labor-intensive. 
This challenge is further exacerbated by the ever-increasing volume of medical image data, making it difficult for radiologists to meet the demands for timely and accurate reporting. 
%The increasing volume of medical imaging data has further exacerbated the challenges, making it difficult for radiologists to keep up with the demand for timely and accurate reports. 
There has been a growing interest in developing automatic report generation methods that can alleviate the burden on medical professionals while maintaining the high standards required in clinical settings.

Numerous approaches have been introduced for 
automatic radiology report generation \cite{chen2020generating, chen2021cross, qin2022reinforced}. 
Most existing studies rely on Transformer models \cite{vaswani2017attention} that have demonstrated impressive performance in a variety of natural language processing tasks such as image captioning and text generation. Transformers leverage self-attention mechanisms to model long-range dependencies, making them particularly well-suited for generating coherent and contextually relevant reports from complex medical images.  
However, 
Transformer models are often criticized for their high computational complexity, limiting their use in real applications.
Recently, the Mamba model~\cite{gu2023Mamba}, designed to reduce computational complexity without compromising performance, has attracted increasing attention. 
Mamba's efficient sequence processing capabilities make it an attractive alternative to Transformers, but its potential for radiology report generation has not yet been fully explored. 
\if false
To overcome these limitations, we explore the Mamba model~\cite{gu2023Mamba}, designed to reduce computational complexity without compromising performance. 
Mamba's efficient sequence processing capabilities make it an attractive alternative, yet its potential in radiology report generation has been under-explored. 
\fi

In this work, we propose a novel radiology report generation method, called R2Gen-Mamba, which leverages the strengths of both Mamba and Transformer architectures. 
%a novel framework that synergizes the strengths of Mamba and Transformer architectures. 
Specifically, R2Gen-Mamba leverages Mamba with low computational complexity as the encoder, and Transformer as the decoder retaining powerful contextual processing capability. 
By combining these complementary models, R2Gen-Mamba provides a new pathway for reducing the computational burden in radiology while ensuring high-quality, contextually relevant reports. 
%此处应该简要介绍一下方法的主要组成，而不是只说方法的优点
Experimental results on two benchmark datasets IU X-Ray~\cite{demner2016preparing} and MIMIC-CXR~\cite{johnson2019mimic}, %  
suggests that R2Gen-Mamba outperforms traditional Transformer-based models 
regarding report quality and computational efficiency. 
Compared with  
state-of-the-art (SOTA) studies, R2Gen-Mamba provides a more resource-efficient solution for automatic radiology report generation.

\section{METHODOLOGY}
\begin{figure}[!t]
    \centering
    \includegraphics[width=\linewidth]{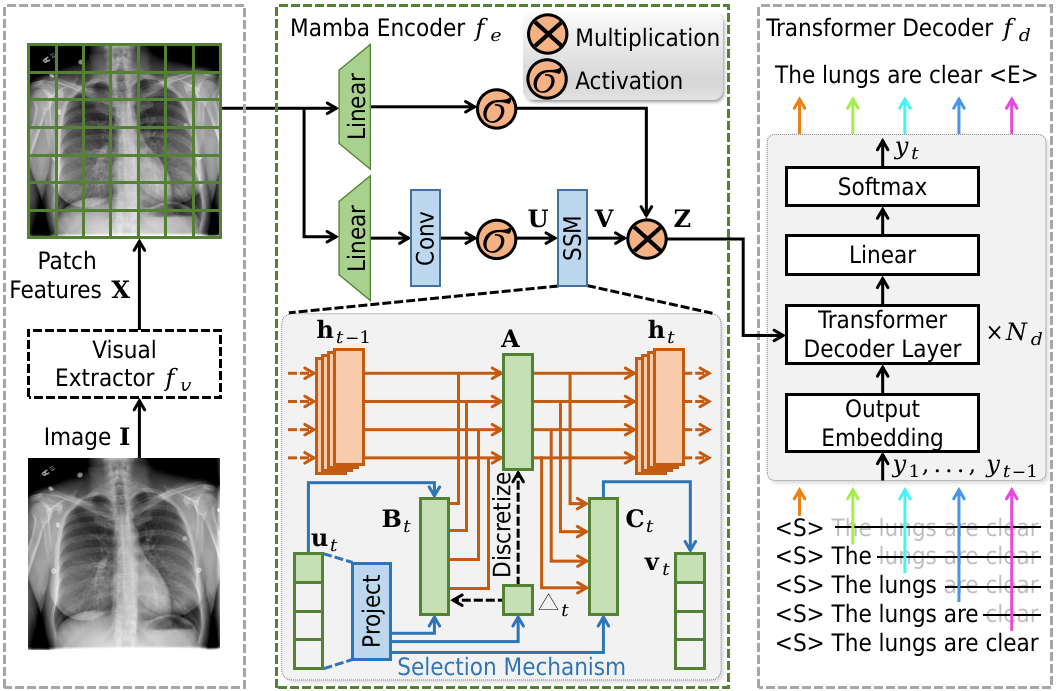} % 图片路径
    \caption{Architecture of the proposed R2Gen-Mamba framework, with visual extractor and decoder denoted by gray dashed boxes. The Mamba encoder is highlighted within green dashed boxes. Conv: convolution; SSM: selective state space model; Linear: linear projection.}
    \label{fig:R2Gen-Mamba}
\end{figure}

Radiology report generation can be framed as a sequence-to-sequence problem, where the input image patch features serve as the input sequence and the corresponding report as the target sequence. Typically, the input patch feature sequence $\mathbf{X}=\{\mathbf{x}_1, \mathbf{x}_2,\dots,\mathbf{x}_S\}$, where $S$ is the number of patches, each $\mathbf{x}_s \in \mathbb{R}^d$, consists of visual features extracted from the image patches using pre-trained visual extractor like convolutional neural networks. The output sequence $Y=\{y_1,y_2,\dots,y_T\}$, where $T$ is the maximum length of reports,  each $y_t$ is a token from a predefined vocabulary, represents the generated report. This sequence-to-sequence framework is optimized through maximum likelihood of generating the correct report given the input image. Our R2Gen-Mamba contains three major parts (\ie, visual extractor, Mamba encoder, and Transformer decoder), which are outlined in subsequent subsections.

\subsection{Visual Extractor}
To produce radiology reports, we begin by extracting visual features from the radiology images using convolutional neural networks such as VGG or ResNet.
As illustrated in Fig.~\ref{fig:R2Gen-Mamba}, the image is passed through the Visual Extractor to extract the feature map. Each spatial pixel in the feature map corresponds to a patch in the original image. These spatial pixels are flattened to obtain a sequence representation that serves as the input sequence for subsequent Mamba encoder. This process is formally represented as: $\{\mathbf{x}_1, \mathbf{x}_2,\dots,\mathbf{x}_S\}=f_v(Img)$,
where $f_v(\cdot)$ is the visual extractor, and $Img$ is the input image. 
% The sequence of visual features thus obtained is utilized as the input for the Mamba encoder model to extract contextual semantic information.

\subsection{Mamba Encoder}
To extract contextual semantic information, we use Mamba as the encoder.  
Mamba is designed to process sequence data. 
Compared with Transformers that have quadratic computational complexity, % of Transformer, 
Mamba has linear complexity for the number of tokens.  Provided the input sequence $\{\mathbf{x}_1, \mathbf{x}_2,\dots,\mathbf{x}_S\}$, the output sequence $\mathbf{Z}$ is obtained by $\{\mathbf{z}_1, \mathbf{z}_2,\dots,\mathbf{z}_S\} = f_e(\mathbf{x}_1, \mathbf{x}_2,\dots,\mathbf{x}_S)$,
% \begin{equation}
%     {z_1, z_2, \dots, z_S} = f_e(x_1, x_2, \dots, x_S)
% \end{equation}
where $f_e$ denotes the Mamba encoder.
As for the core state space model (SSM) of Mamba, given the input sequence $\mathbf{U}$, the output sequence $\mathbf{V}$ is obtained by $\{\mathbf{v}_1, \mathbf{v}_2,\dots,\mathbf{v}_S\} = \text{SSM}(\mathbf{u}_1, \mathbf{u}_2,\dots,\mathbf{u}_S)$.
% \begin{equation}
%  {v_1, v_2, \dots, v_S} = \text{SSM}(u_1, u_2, \dots, u_S)   
% \end{equation}
Specifically, as illustrated in Fig. \ref{fig:R2Gen-Mamba}, $\mathbf{u}_t$, $t \in \{1, 2, \dots, S\}$ is fed into linear layers to obtain continuous parameters: $\mathbf{B}_t, \mathbf{C}_t, \Delta_t = \text{Project}(\mathbf{u}_t)$. Then discretization is performed by zero-order hold (ZOH): $\Bar{\mathbf{A}}_t = \exp(\Delta_t \mathbf{A})$; $\Bar{\mathbf{B}}_t = (\Delta_t \mathbf{A})^{-1}(\exp(\Delta_t \mathbf{A}) -\mathbf{I})\cdot \Delta_t \mathbf{B}_t$, where $\mathbf{A}$ is a learnable embedding. Finally, the sequence-to-sequence transformation is achieved in two stages: $\mathbf{h}_t = \Bar{\mathbf{A}}_t \mathbf{h}_{t-1} + \Bar{\mathbf{B}}_t \mathbf{u}_t$; $\mathbf{v}_t = \mathbf{C}_t \mathbf{h}_t$.

% \begin{equation}
%     B_t, C_t, \Delta_t = \text{Project}(u_t)
% \end{equation}
% \begin{equation}
%     \Bar{A_t} = \exp(\Delta_t A)
% \end{equation}
% \begin{equation}
%     \Bar{B_t} = (\Delta_t A)^{-1}(\exp(\Delta_t A) -I)\cdot \Delta_t B_t
% \end{equation}

% \begin{equation}
%     h_t = \Bar{A_t} h_{t-1} + \Bar{B_t} u_t
% \end{equation}
% \begin{equation}
%     v_t = C h_t
% \end{equation}
% where $A$ is learnable parameters.

\subsection{Transformer Decoder}
In the proposed R2Gen-Mamba, the decoder is built upon the standard Transformer architecture. 
%As illustrated in Fig.~\ref{fig:R2Gen-Mamba}, '=
The decoding procedure is formulated as: 
$y_t = f_d(\mathbf{z}_1, \mathbf{z}_2,\dots,\mathbf{z}_S, y_1, \dots,  y_{t-1})$,
where $f_d(\cdot)$ is the Transformer decoder. 
As noted in \cite{vaswani2017attention}, the decoder needs to rely on the generation results of the previous step due to its auto-regressive nature and requires additional attention mechanisms, so we repeat the decoder layer $N_d$ times.  In our experiments, we  set $N_d$ to 3.

\subsection{Objective Function}
The overall generation process in R2Gen-Mamba can be mathematically framed as a recursive implementation of the chain rule, where the probability of the target sequence $\{y_1,y_2,\dots,y_T\}$ provided the input image $Img$ is expressed as: 
$p(Y \mid Img) = \prod\limits_{t=1}^T p(y_t \mid y_1, \dots, y_{t-1}, Img)$. 
The model is trained by maximizing the likelihood of the target sequence conditioned on the input image:
 \begin{equation}
     \theta^* = \operatorname*{argmax}_{\theta} \sum\nolimits_{t=1}^T \log p(y_t \mid y_1, \dots, y_{t-1}, Img; \theta)
 \end{equation}
where $\theta^*$ represents the parameters of R2Gen-Mamba. This optimization process ensures that the model learns to accurately generate the report text based on the visual features extracted from the input image. 
During inference, we use the beam search strategy to sample predictions. 
To facilitate reproducible research, we have shared the source code to the public through  \href{https://github.com/mxliu/ACTION-Software-for-Functional-MRI-Analysis/tree/main/Software}{GitHub}.

\section{EXPERIMENTS}

% \usepackage{booktabs}
% \usepackage{multirow}
\begin{table}[]
\setlength{\abovecaptionskip}{0pt}
\setlength{\belowcaptionskip}{0pt}
\setlength{\abovedisplayskip}{0pt}
\setlength{\belowdisplayskip}{0pt}
\small
\renewcommand{\arraystretch}{0.9}
\caption{Details of two benchmark datasets used in this work.}
\label{tab:dataset}
\resizebox{\columnwidth}{!}{%
\begin{tabular}{@{}l|lll|lll@{}}
\toprule
\multirow{2}{*}{Dataset} & \multicolumn{3}{c|}{IU X-Ray} & \multicolumn{3}{c}{MIMIC-CXR} \\ \cmidrule(l){2-7} 
                         & Train    & Validation      & Test     & Train     & Validation     & Test    \\ \midrule
Image \#                 & 5.23K    & 0.75K      & 1.50K    & 368.96K   & 2.99K   & 5.16K   \\
Report \#                & 2.77K    & 0.40K      & 0.79K      & 222.76K   & 1.81K   & 3.27K   \\
Patient \#               & 2.77K    & 0.40K      & 0.79K      & 64.59K    & 0.50K     & 0.29K     \\
Average Length              & 37.56    & 36.78    & 33.62    & 53.00     & 53.05   & 66.40   \\ \bottomrule
\end{tabular}}
\end{table}

% \usepackage{booktabs}
% \usepackage{multirow}
\begin{table*}[]
\setlength{\abovecaptionskip}{0pt}
\setlength{\belowcaptionskip}{2pt}
\setlength{\abovedisplayskip}{0pt}
\setlength{\belowdisplayskip}{0pt}
\footnotesize
\centering
\setlength{\tabcolsep}{6.5pt}
\caption{Comparisons of different methods on IU X-Ray and MIMIC-CXR. `BLEU-x': BLEU score with an n-gram size of x. The best results are highlighted in bold.}
\label{tab:R2Gen-Mamba}
\begin{tabular}{@{}l|c|llllll|ccc@{}}
\toprule
\multirow{2}{*}{Data}      & \multirow{2}{*}{Method} & \multicolumn{6}{c|}{NLG Metrics}                                                                                                                                & \multicolumn{3}{c}{CE Metrics}                                                    \\ \cmidrule(l){3-11}
% \cline{3-11}
                           &                        & \multicolumn{1}{c}{BLEU-1} & \multicolumn{1}{c}{BLEU-2} & \multicolumn{1}{c}{BLEU-3} & \multicolumn{1}{c}{BLEU-4} & \multicolumn{1}{c}{METEOR} & \multicolumn{1}{c|}{ROUGE-L} & Precision                         & Recall                         & F1 score                       \\ \midrule
\multirow{4}{*}{IU X-Ray}  & R2Gen                  & 0.423                    & 0.275                    & 0.203                    & 0.160                    & 0.176                   & 0.358                     & -                         & -                         & -                         \\
                           & R2Gen-CMN              & 0.470                    & 0.300                    & 0.215                    & 0.166                    & 0.189                   & 0.367                     & -                         & -                         & -                         \\
                           & R2Gen-RL               & 0.291                    & 0.178                    & 0.121                    & 0.086                    & 0.096                   & 0.312                     & -                         & -                         & -                         \\ \cmidrule(l){2-11} 
                           & R2Gen-Mamba (Ours)                  & \textbf{0.482}           & \textbf{0.315}           & \textbf{0.228}           & \textbf{0.176}           & \textbf{0.208}          & \textbf{0.382}            & -                         & -                         & -                         \\ \midrule
\multirow{4}{*}{MIMIC-CXR} & R2Gen                  & \textbf{0.371}                    & \textbf{0.223}                    & 0.148                    & 0.105                    & 0.141                   & 0.271                     & {0.429} & {0.243} & 0.310 \\
                           & R2Gen-CMN              & 0.352           & 0.214                    & 0.141                    & 0.099                    & 0.139          & 0.274                     & {0.441} & \textbf{0.326} & 0.375 \\
                           & R2Gen-RL               & 0.122                    & 0.067                    & 0.042                    & 0.028                    & 0.047                   & 0.137                     & 0.061
                         & 0.027
                          & 0.038                          \\ \cmidrule(l){2-11} 
                           & R2Gen-Mamba (Ours)                   & 0.352                    & 0.222           & \textbf{0.152}           & \textbf{0.110}           & \textbf{0.141}                   & \textbf{0.284}            & \textbf{0.483}                          & 0.325                          & \textbf{0.389}                          \\ \bottomrule
\end{tabular}
\end{table*}

\subsection{Experimental Setup}
We perform experiments on two benchmark datasets: IU X-Ray~\cite{demner2016preparing} and MIMIC-CXR~\cite{johnson2019mimic}. The IU X-Ray dataset includes 7,470 chest X-ray images paired with 3,955 reports, while MIMIC-CXR comprises 473,057 images and 206,563 reports. Following prior studies~\cite{chen2020generating, chen2021cross, qin2022reinforced}, we exclude samples without reports. We use a 70\%/10\%/20\% split for training, validation, and testing on IU X-Ray, and the official split for MIMIC-CXR, as detailed in Table~\ref{tab:dataset}. Two evaluation metrics are employed: traditional natural language generation (NLG) metrics (BLEU~\cite{papineni2002bleu}, METEOR~\cite{denkowski2011meteor}, and ROUGE-L~\cite{lin2004rouge}) and clinical efficacy (CE) metrics. For CE metrics, we use the CheXbert~\cite{smit2020combining} tool to automatically label generated reports, comparing them to ground truths across 14 thoracic disease categories using precision, recall, and F1 score.

\subsection{Implementation Details}
Following~\cite{chen2020generating, chen2021cross, qin2022reinforced}, we use two images per patient for IU X-Ray and one image for MIMIC-CXR as input. 
The visual extractor utilizes a ResNet101 model pre-trained on ImageNet, with patch features projected to a dimension of 512. The Mamba encoder is set to a dimension of 512, with an SSM state expansion factor of 16, a local convolution width of 4, and a block expansion factor of 2. 
The Transformer decoder also has a dimension of 512, with 3 layers, 8 heads, and a dropout rate of 0.1. 
We use the Adam optimizer and set learning rates of $5 \times 10^{-5}$ for the visual extractor and $1 \times 10^{-4}$ for other parameters, decayed by 0.8 per epoch. 
The model that achieved the best BLEU-4 score on the validation sets is selected, with a beam size of 3 for inference to balance between generation quality and computational efficiency. 

%\section{RESULTS}

\begin{figure*}[h]
\setlength{\abovecaptionskip}{2pt}
\setlength{\belowcaptionskip}{0pt}
\setlength{\abovedisplayskip}{0pt}
\setlength{\belowdisplayskip}{0pt}
    \centering
    \includegraphics[width=1\linewidth]{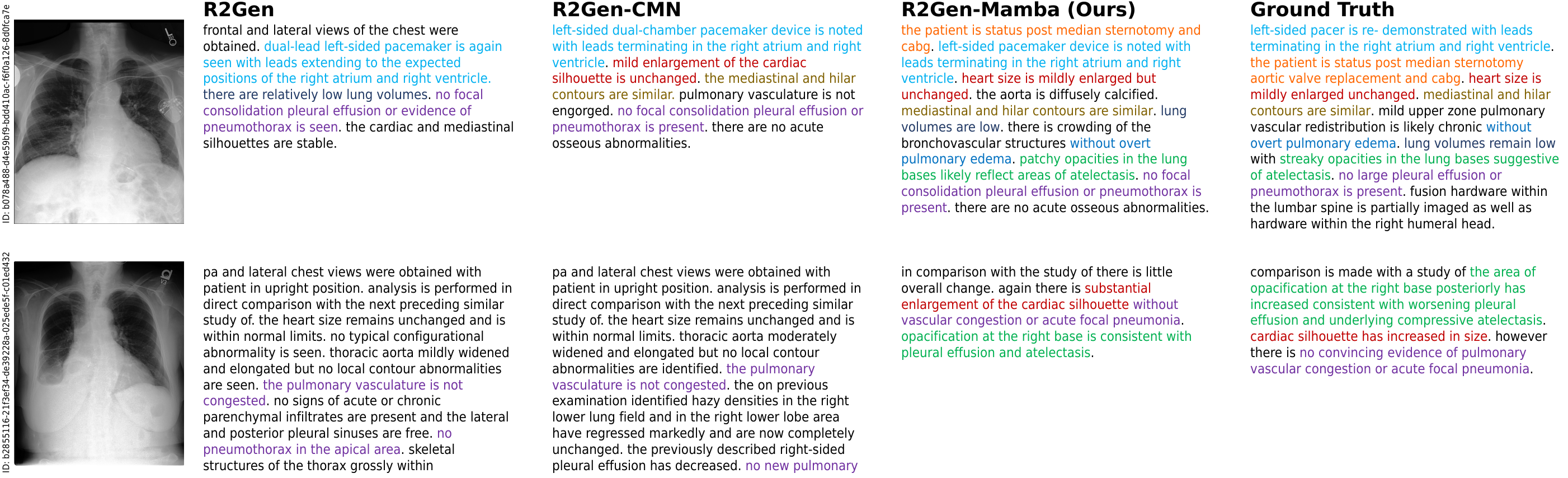} % 图片路径
    \caption{Examples of ground truth and generated reports by different methods, with similar findings marked in the same color.}
    \label{fig:example}
\end{figure*}

\subsection{Visual and Quantitative Results}
To evaluate the effectiveness of our R2Gen-Mamba, we performed a comparative analysis against existing SOTA methods, namely R2Gen~\cite{chen2020generating}, R2Gen-CMN~\cite{chen2021cross}, and R2Gen-RL~\cite{qin2022reinforced}. 
Using the same data,  
R2Gen and R2Gen-CMN were implemented using their released code and checkpoints for inference, and R2Gen-RL was retrained from scratch using their released code.
Several typical reports generated by different methods are shown in Fig.~\ref{fig:example}. 
It can be seen from this figure that the report generated by R2Gen-Mamba contains more precise information, 
providing superior results than the competing methods in accuracy and clarity. 
The quantitative results regarding NLG and CE metrics are summarized in Table \ref{tab:R2Gen-Mamba}, from which we have  
%Our analysis reveals 
several key findings.

\emph{Firstly}, our R2Gen-Mamba, which incorporates Mamba and Transformer, outperforms existing approaches in most cases, suggesting 
the advantages of Mamba for report generation and the feasibility of combining Mamba with Transformer. 
\emph{Secondly}, R2Gen-Mamba slightly under-performs R2Gen on BLEU-1 and BLEU-2 metrics for MIMIC-CXR but surpasses it on BLEU-3, BLEU-4, METEOR, and ROUGE-L. 
BLEU-1 and BLEU-2 measure the overlap of single words and word pairs, reflecting basic vocabulary matching. BLEU-3 and BLEU-4 measure triples and quadruples, capturing longer context dependencies. Higher BLEU-3 and BLEU-4 scores indicate R2Gen-Mamba generates text with better grammatical and semantic structures, reflecting stronger context modeling and grammatical consistency. METEOR combines lexical matching, word order, and morphological changes, while ROUGE-L assesses the longest common subsequence between generated and reference texts. Our R2Gen-Mamba's better performance on these metrics demonstrates stronger vocabulary choice, grammatical structure, and alignment with reference text.
\emph{Thirdly}, R2Gen-Mamba demonstrates superior performance on clinical efficacy (CE) metrics, suggesting that the generated reports offer more valuable clinical information for diagnosis and decision-making. 
This highlights the clinical relevance and utility of our R2Gen-Mamba compared with the competing methods.

\subsection{Computation Complexity Analysis}
With the Mamba encoder in the proposed R2Gen-Mamba framework, we can significantly reduce model complexity, with only 594.944~K parameters and incurring a computational load of 58.216~M floating-point operations (FLOPs). This represents a substantial improvement over the Transformer encoder utilized in the SOTA R2Gen model, which comprises 4.728~M parameters and incurs a computational complexity of 462.422~M FLOPs. 
The considerable reduction in both parameter count and computational cost highlights the efficiency of the Mamba encoder, making it more suitable for resource-constrained environments while maintaining superior performance in radiology report generation.

\section{CONCLUSION}
This paper presents R2Gen-Mamba, a novel radiology report generation model that leverages Mamba's efficient sequence processing and Transformer's contextual strengths. 
R2Gen-Mamba reduces computational complexity while producing high-quality radiology reports.
Experiments on two datasets show that R2Gen-Mamba surpasses existing methods in both natural language generation and clinical efficacy metrics. 
Our findings highlight the effectiveness of merging Mamba with Transformer techniques for radiology report generation. 
%R2Gen-Mamba not only meets but often exceeds state-of-the-art performance, offering a promising solution for automatic report generation. 
%Future work will focus on refining and expanding the framework's clinical applications.

\section{COMPLIANCE WITH ETHICAL STANDARDS}
This research was conducted retrospectively using human subject data made available in open access by IU X-Ray and MIMIC-CXR.
Ethical approval was not required as confirmed by the license attached with the open-access data.

\section{ACKNOWLEDGMENTS}
%This work was partly supported by NIH grant (No.~AG073297). 
The research of M.~Liu and H.~Zhu was supported in part by NIH grants AG073297 and AG082938. %, EB035160, and NS134849.  
The research of C.~Lian was supported in by NSFC Grants (Nos. 12326616, 62101431, and 62101430) and Natural Science Basic Research Program of Shaanxi (No. 2024JC-TBZC-09). 
%This work was finished when Y.~Sun visited UNC-Chapel Hill. 
%{\color{red}Data used in this paper were obtained from the Alzheimer's Disease Neuroimaging Initiative (ADNI).}
%\vfill
%\pagebreak

%\section{Acknowledgments}
%\label{sec:acknowledgments}


% -------------------------------------------------------------------------
%\small
\bibliographystyle{IEEEbib}
\bibliography{R2Gen-Mamba_simple2}

\end{document}